\pdfoutput=1

\documentclass[11pt]{article}

\usepackage{ACL2023}
\usepackage{times}
\usepackage{latexsym}
\usepackage[T1]{fontenc}
\usepackage[utf8]{inputenc}
\usepackage{microtype}
\usepackage{inconsolata}
\usepackage{graphicx}
\usepackage{tikz}
\usepackage{booktabs}
\usepackage{multirow}
\usepackage{amsmath}
\DeclareUnicodeCharacter{2032}{'}
\DeclareUnicodeCharacter{2033}{''}

\title{Table and Image Generation for Investigating Knowledge of Entities in Pre-trained Vision and Language Models}

\author{
  Hidetaka Kamigaito\textsuperscript{\dag}, Katsuhiko Hayashi\textsuperscript{\ddag}, Taro Watanabe\textsuperscript{\dag} \\
  \textsuperscript{\dag}Nara Institute of Science and Technology \textsuperscript{\ddag}Hokkaido University \\
  \texttt{\{kamigaito.h, taro\}@is.naist.jp} \\
  \texttt{katsuhiko-h@ist.hokudai.ac.jp}}

\begin{document}
\maketitle
\begin{abstract}

In this paper, we propose a table and image generation task to verify how the knowledge about entities acquired from natural language is retained in Vision \& Language (V\&L) models. This task consists of two parts: the first is to generate a table containing knowledge about an entity and its related image, and the second is to generate an image from an entity with a caption and a table containing related knowledge of the entity. In both tasks, the model must know the entities used to perform the generation properly. We created the Wikipedia Table and Image Generation (WikiTIG) dataset from about 200,000 infoboxes in English Wikipedia articles to perform the proposed tasks. We evaluated the performance on the tasks with respect to the above research question using the V\&L model OFA \cite{pmlr-v162-wang22al}, which has achieved state-of-the-art results in multiple tasks. Experimental results show that OFA forgets part of its entity knowledge by pre-training as a complement to improve the performance of image related tasks.
\end{abstract}

\section{Introduction}

Vision \& Language (V\&L), which is the fusion of vision and language tasks, has achieved great success in tasks such as caption generation from images \cite{pmlr-v37-xuc15} and image generation from texts \cite{pmlr-v48-reed16}. This progress has been driven by pre-trained V\&L models that are trained on large-scale V\&L datasets \cite{ijcai2022p0762}. To generate appropriate captions and images for input, pre-trained V\&L models need to have prior knowledge of the features of the objects they are generating \cite{10.1007/978-3-030-58539-6_34,yun-etal-2021-vision-language}. These models retain knowledge about entities in particular by inheriting parameters from pre-trained language models used in natural language processing to indirectly utilize data resources such as Wikipedia.

In this way, V\&L models \cite{NEURIPS2019_c74d97b0,Su2020VL-BERT:,Li_Duan_Fang_Gong_Jiang_2020,pmlr-v139-cho21a,pmlr-v162-wang22al,saharia2022photorealistic} map the inherited textual knowledge into visual representations through additional training on V\&L datasets.

This learning process raises a number of questions, such as whether the knowledge about entities acquired from natural language is adequately retained in the pre-trained V\&L model, or whether it is enhanced by combining it with image features. These are important in understanding the limits of what can be generated by the pre-trained V\&L model.

\begin{figure}
\centering
\includegraphics[width=\columnwidth]{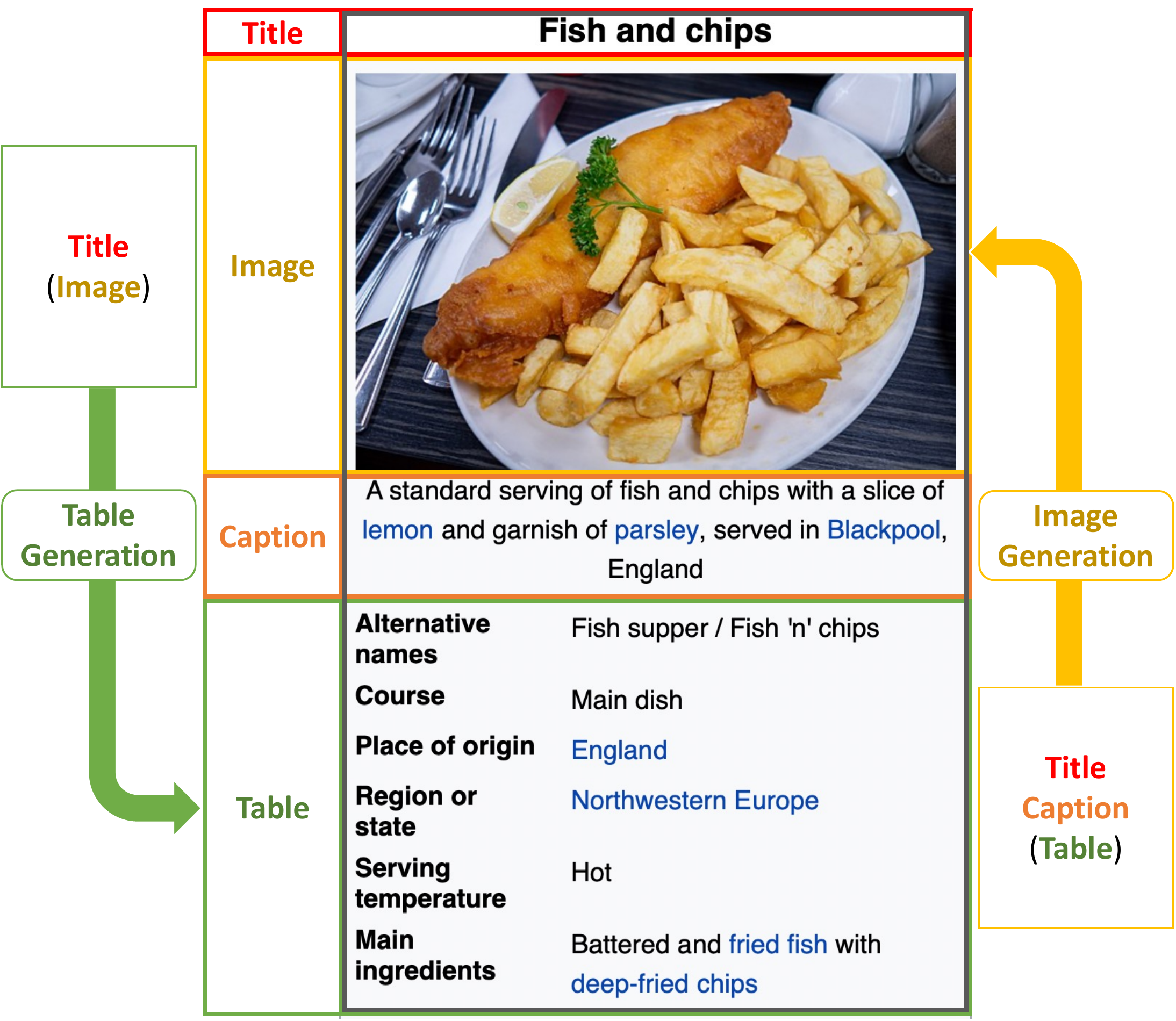}
\caption{An infobox of a Wikipedia article\footnotemark. In this study, we validate the V\&L model by generating images and tables in infoboxes.}
\label{fig:infobox}
\end{figure}

\footnotetext{\url{https://en.wikipedia.org/wiki/Fish_and_chips}}

To answer these questions, we propose a task of generating tables and images of infoboxes in English Wikipedia. Figure \ref{fig:infobox} shows an example of the target infobox, in which either tables or images are generated by the proposed task. In both cases, the model must know the entities to generate them properly.

We collected about 200,000 infoboxes to construct the Wikipedia Table and Image Generation (WikiTIG) dataset necessary to perform the proposed task. In addition, we used OFA \cite{pmlr-v162-wang22al}, a pre-trained V\&L model that has achieved state-of-the-art performance in various V\&L tasks.

Our evaluation of the table generation revealed that part of the knowledge in the V\&L model acquired from natural language is lost when the V\&L model is pre-trained. We also found that additional knowledge for entities was acquired by supplementing image information, which was not possible solely from textual data.

In image generation, we found that OFA can generate more accurate images by using the knowledge expressed in the table. We also found that the models trained only on natural language can infer table knowledge, which increases the diversity of generated images. Our code and dataset will be released at \url{https://github.com/kamigaito/WikiTIG}.

\begin{figure}
    \centering
    \includegraphics[width=\columnwidth]{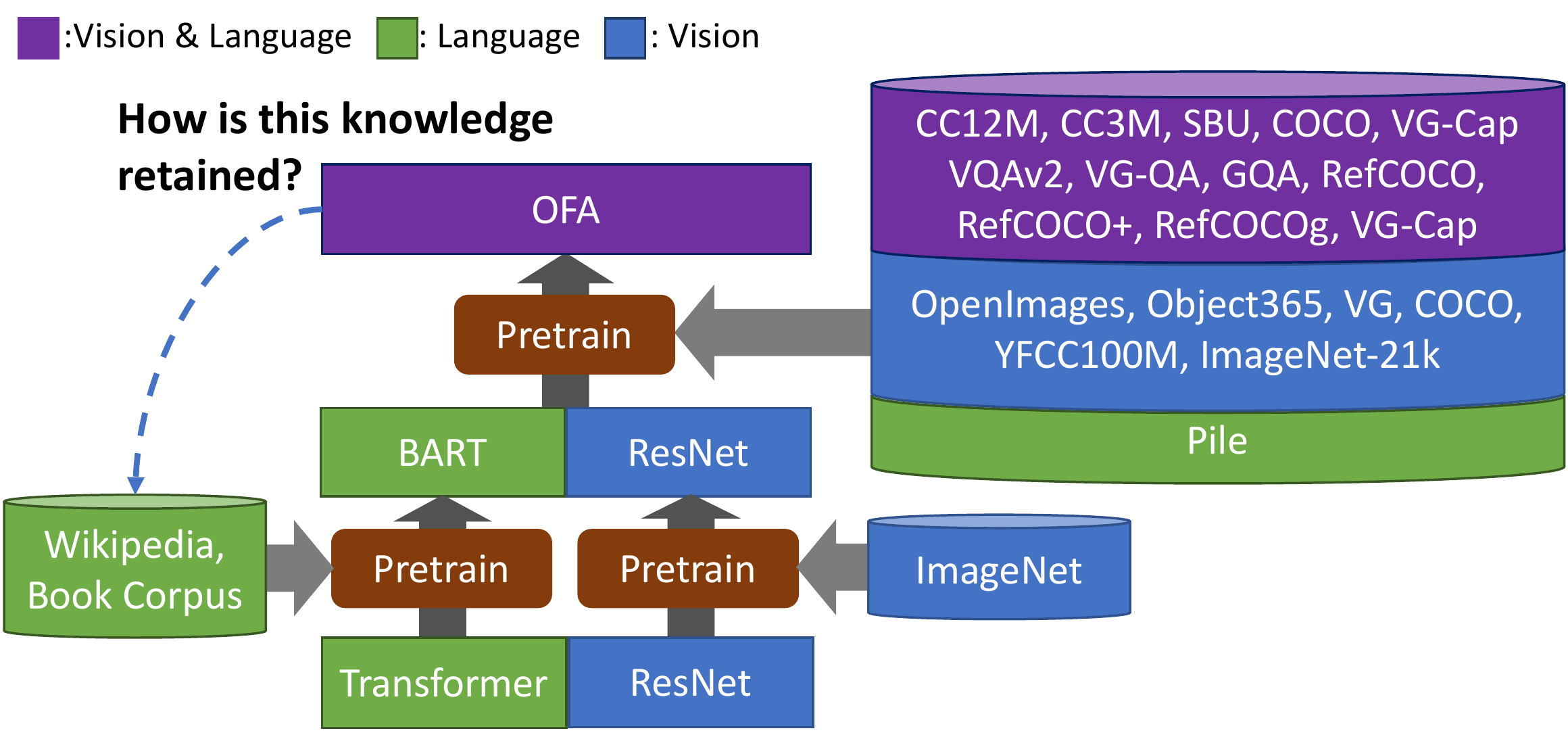}
    \caption{Learning process of OFA. We investigate how OFA retains knowledge about entities acquired from pre-training on Wikipedia articles.}
    \label{fig:overview}
\end{figure}

\section{Vision \& Language Models}

Many pre-trained V\&L models have achieved state-of-the-art performance on various tasks by inheriting the weights of the conventional pre-trained models for natural language and images \cite{NEURIPS2019_c74d97b0,Su2020VL-BERT:,Li_Duan_Fang_Gong_Jiang_2020,pmlr-v139-cho21a,pmlr-v162-wang22al,saharia2022photorealistic} before learning V\&L datasets. Our study examines how the knowledge represented in the pre-trained model for natural language is transformed through such a learning process. We select OFA, which has achieved state-of-the-art performance in multiple V\&L tasks, as our target model.

Figure \ref{fig:overview} shows the network structure of OFA and its relation to each dataset\footnote{Appendix \ref{appendix:ofa} describes the data for the pre-training.}. OFA uses VQGAN \cite{esser2020taming} on the decoder to transform images into discrete sequences so that the same Transformer \cite{NIPS2017_3f5ee243} is used for image and natural language generation. Because OFA inherits parameters from BART \cite{lewis-etal-2020-bart}, which shares a similar Transformer structure, OFA should include knowledge acquired from natural language such as Wikipedia articles. Unlike the decoder, the encoder handles images directly; thus, OFA uses the output of ResNet \cite{He_2016_CVPR} to embed images in addition to the embedding layer inherited from BART.

\section{Table and Image Generation}

In this section, we describe two tasks for verifying knowledge behavior in the V\&L model: table generation and image generation. Both tasks are based on infoboxes in Wikipedia articles, which correspond to summary information of the Wikipedia articles comprising tables and images\footnote{\url{https://en.wikipedia.org/wiki/Help:Infobox}}. Thus, it is suitable for verifying the knowledge about entities in Wikipedia kept in the pre-trained V\&L model. In the following subsections, we explain the details of each task.

\begin{table}[t]
    \centering
    \small
    \begin{tabular}{lcc}
        \toprule
         \textbf{Task} & \textbf{Input} & \textbf{Output} \\
         \midrule
         Table Generation & Title, Image & Table \\
         Image Generation & Title, Caption, Table & Image \\
         \bottomrule
    \end{tabular}
    \caption{Outline of each task. See Figure \ref{fig:infobox} for the parts of the infobox to which each term refers.}
    \label{tab:tasks}
\end{table}
\begin{figure}[t]
    \centering
    \begin{tikzpicture}
    \node[text width=0.95\columnwidth,draw] at (0,0) 
    {\scriptsize Alternative names \verb+|+ Fish supper / Fish 'n' chips \verb|<|\verb|>| Course \verb+|+ Main dish \verb|<|\verb|>| Place of origin \verb+|+ England \verb|<|\verb|>| Region or state \verb+|+ Northwestern Europe \verb|<|\verb|>| Serving temperature \verb+|+ Hot \verb|<|\verb|>| Main ingredients \verb+|+ Battered and fried fish with deep-fried chips};
    \end{tikzpicture}
    \caption{This example is a linearized version of the table in Figure \ref{fig:infobox}.}
    \label{fig:linearized}
\end{figure}

\subsection{Table Generation}
\label{subsec:tab_gen}

In the table generation task, the target V\&L model generates a table from a title and/or image of the infobox. To do this, the model generates linearized tables, similarly to table generation by descriptions \cite{wu-etal-2022-text-table}. In our setting, we linearize tables as shown in Figure \ref{fig:linearized} using the column separator ``\verb+|+'' and the row separator ``\verb|<|\verb|>|'' to reuse pre-trained token embeddings. The separator symbols are accompanied by spaces before and after for use in BPE tokenization. We investigate the target model by directly generating such linearized text. We use the following settings for the investigation.

\paragraph{Generation from titles} We investigate the knowledge about entities held by V\&L models by comparing tables generated from titles by pre-trained V\&L models and by pre-trained models trained only on natural language.

\paragraph{Generation from title and images} We generate tables from titles with images and compare the results with those generated from only titles. This enables us to investigate the new knowledge in pre-trained V\&L models transferred from images.

\paragraph{Metrics} For comparison, we use the following evaluation metrics to measure how close the generated tables are to the actual ones.

\noindent - \textbf{ROUGE}: Since the linearized tables are text data and the infobox plays the role of summarizing the article, we use ROUGE \cite{lin-2004-rouge}, the most widely used evaluation method for automatic summarization. In our evaluation with ROUGE, we convert the column separator ``\verb+|+'' and the row separator ``\verb|<|\verb|>|'' to spaces so that the sequence of strings is not restricted to rows and columns.

\noindent - \textbf{Table-F$_1$}: To evaluate the tables with respect to their structure, we divide the cells by their types and then evaluate the matches with the reference table in terms of the F$_1$ measure for each case and average them. When calculating the matches, we apply clipping used in ROUGE to prevent the score from increasing due to the repetition of the same cell in the output\footnote{Appendix \ref{appendix:metric:tf1} shows the details of this calculation.}. We treat cells of each type separately\footnote{Appendix \ref{appendix:group_names} shows an example of the cell types.} as follows:
\begin{itemize}

\item\textbf{Group}: The infobox sometimes divides the table into groups, with the first row of each group serving as a header for the group name. The prediction performance for the group names is important for verifying what aspects of knowledge the model has about the entities. Since these rows consist of a single column, we target rows consisting of a single column in this type of cell.

\item\textbf{Header}: The head of each row in the table consisting of more than one column is usually the header of a subsequent cell in the same row. Therefore, the prediction performance for headers is important for the same reason as for group names.

\item\textbf{Value}: The second cells in each row of a table with two columns have values corresponding to the headers. Therefore, the prediction performance of the values is important for knowing whether the model has detailed knowledge about the entity. To examine the correspondence between headers and their values, we treat a header and its corresponding value as a pair.

\end{itemize}

\noindent - \textbf{Corpus-F$_1$}: Because the above Table-F$_1$ computes each case individually, it is difficult to evaluate how much diverse knowledge the model outputs. To solve this problem, we share cells across all instances and compute F$_1$ values in a batch. Similarly to Table-F$_1$, we apply clipping to the score calculation\footnote{Appendix \ref{appendix:metric:cf1} shows the details of this calculation.} and treat cell types Group, Header, and Value separately as defined in Table-F$_1$.

\begin{table}[t]
    \centering
    \small
    \resizebox{\columnwidth}{!}{
    \begin{tabular}{lcccc}
        \toprule
        \textbf{Task} & \textbf{Total} & \textbf{Train} & \textbf{Valid} & \textbf{Test} \\
         \midrule
         Table Generation & 204,460 & 184,124 & 10,081 & 10,255 \\
         Image Generation & 86,654 & 78,012 & 4,261 & 4,381 \\
        \bottomrule
    \end{tabular}}
    \caption{The data size for each task in the WikiTIG dataset.}
    \label{tab:stats}
\end{table}
\begin{table*}[t]
\centering
\resizebox{\textwidth}{!}{
\begin{tabular}{llccccccccc}
\toprule
\multirow{2}{*}{\textbf{Model}} & \multirow{2}{*}{\textbf{Input}} & \multicolumn{3}{c}{\textbf{ROUGE}~$\uparrow$} & \multicolumn{3}{c}{\textbf{Table-F$_1$}~$\uparrow$} & \multicolumn{3}{c}{\textbf{Corpus-F$_1$}~$\uparrow$} \\
\cmidrule(lr){3-5}\cmidrule(lr){6-8}\cmidrule(lr){9-11}
                       &                        & \textbf{1}       & \textbf{2}      & \textbf{L}      & \textbf{Header}   & \textbf{Group}   & \textbf{Value}  & \textbf{Header}   & \textbf{Group}   & \textbf{Value}  \\
                       \midrule
BART                   & Title & 28.8${\scriptstyle\pm0.2}$ & 14.0${\scriptstyle\pm0.1}$ & 26.6${\scriptstyle\pm0.1}$ & 38.9${\scriptstyle\pm0.1}$ & \underline{\textbf{24.3}}${\scriptstyle\pm0.1}$ & \underline{\textbf{4.9}}${\scriptstyle\pm0.0}$ & \underline{\textbf{62.9}}${\scriptstyle\pm0.3}$ & \underline{\textbf{35.5}}${\scriptstyle\pm0.0}$ & \underline{\textbf{11.7}}${\scriptstyle\pm0.0}$ \\
OFA                    & Title & 28.1${\scriptstyle\pm0.2}$ & 13.4${\scriptstyle\pm0.1}$ & 25.7${\scriptstyle\pm 0.2}$ & 34.7${\scriptstyle\pm0.4}$ & 22.8${\scriptstyle\pm0.2}$ & 4.3${\scriptstyle\pm0.1}$ & 57.8${\scriptstyle\pm0.7}$ & 33.3${\scriptstyle\pm0.2}$ & 10.7${\scriptstyle\pm0.2}$ \\
OFA                    & Image & 28.0${\scriptstyle\pm0.1}$ & 11.5${\scriptstyle\pm0.0}$ & 25.8${\scriptstyle\pm0.1}$ & 41.9${\scriptstyle\pm0.1}$ & 21.2${\scriptstyle\pm0.1}$ & 2.7${\scriptstyle\pm0.0}$ & 57.4${\scriptstyle\pm0.2}$ & 26.6${\scriptstyle\pm0.2}$ & 6.8${\scriptstyle\pm0.0}$ \\
OFA                    & Both & \underline{\textbf{31.3}}${\scriptstyle\pm0.1}$ & \underline{\textbf{14.2}}${\scriptstyle\pm0.1}$ & \underline{\textbf{28.7}}${\scriptstyle\pm0.1}$ & \underline{\textbf{43.5}}${\scriptstyle\pm0.1}$ & 23.2${\scriptstyle\pm0.1}$ & 3.7${\scriptstyle\pm0.0}$ & 59.2${\scriptstyle\pm0.2}$ & 28.6${\scriptstyle\pm0.1}$ & 8.2${\scriptstyle\pm0.1}$\\
\bottomrule
\end{tabular}}
\caption{Table generation results. Bold font denotes the highest score, and $\uparrow$ denotes that the higher the score, the more optimal. $\pm$ denotes the standard deviation of the score. \textit{Both} means the input contains both a title and image. Underline indicates that the score improvement is statistically significant from the second-highest one ($p<0.05$)\footnotemark.}
\label{tab:results_table_gen}
\end{table*}

\subsection{Image Generation}

In the image generation task, the model receives a title, caption, and table to generate the corresponding image:

\paragraph{Generation from a title and caption} By using the minimum input required to generate images, we investigate the difficulty of generating them compared to other datasets.

\paragraph{Generation from a title, caption, and table} We investigate the impact of knowledge about entities on image generation by generating images from input, including tables, and compare the results to the setting without tables.

\paragraph{Metrics} We use the following three widely used measures for evaluating image generation.

\noindent - \textbf{CLIP:} The relevance of the input text to the generated images inferred by the pre-trained V\&L model CLIP~\cite{pmlr-v139-radford21a}.

\noindent - \textbf{Inception Score (IS)}: How easily a model can distinguish the differences between each image and the variety of generated images \cite{NIPS2016_8a3363ab}. It is inferred by the pre-trained image classification model Inception-v3~\cite{7780677}.

\noindent - \textbf{Frechet Inception Distance (FID)}: How close the generated image is to the reference image, estimated by Inception-v3 like IS. A lower FID is more ideal.

\footnotetext{We used paired-bootstrap resampling \cite{koehn-2004-statistical} for the significance test.}

\section{Dataset Creation}

We created the Wikipedia Table and Image Generation (WikiTIG) dataset by extracting infoboxes from the HTML dump data of the English Wikipedia\footnote{\url{https://dumps.wikimedia.org/other/static_html_dumps/current/en/} (CC BY-SA 3.0).}. To ensure consistency in the format of infoboxes, we limited the extraction target to those containing a title in the first row and an image in the second row, as shown in Figure \ref{fig:infobox}.

In order to use only entities with sufficient information, we targeted entities for which the table was not empty. In addition, to ensure reliable correspondence, only rows one column wide, which often describe groups, and rows two columns wide, which often consist of a header and its value, were targeted for extraction.

The target images are limited to those in jpeg, png, and gif formats. Since some captions do not include a title, we used a hyphen to join the title at the beginning of the caption in such cases.

Table \ref{tab:stats} shows the size of each dataset. The dataset size diverges between two tasks because some infoboxes do not include captions\footnote{See Appendix \ref{appendix:dataset} for the dataset details.}.

\section{Evaluation \& Analysis}

\subsection{Table Generation}
\label{subsec:exp:table}

\paragraph{Settings} We chose OFA \cite{pmlr-v162-wang22al}, a pre-trained V\&L model, and BART \cite{lewis-etal-2020-bart}, pre-trained only in natural language, as models for comparison. For both models, we used the \texttt{base} settings with the hyperparameters reported in \citet{pmlr-v162-wang22al}. We performed the training three times with different seeds and reported their average scores with their standard deviations\footnote{See Appendix \ref{appendix:exp:table} for the detailed settings.}.

\paragraph{Results} Table \ref{tab:results_table_gen} shows the results for each setting in the table generation\footnote{Appendix \ref{appendix:generated:tables} shows the generated images.}. When only the title is used as input, the result of BART is more accurate than that of OFA, indicating that part of the knowledge acquired from natural language is lost due to additional learning in the V\&L model. The use of image information improves Table-F$_1$ for headers, indicating that images reinforce the knowledge of what kind of features an entity has.

In contrast, F$_1$ for cell values did not improve, indicating that information obtained from images does not complement detailed knowledge, such as the values corresponding to each header obtained from natural language.

The results of BART in Corpus-F$_1$ also suggest that BART contains more diverse knowledge internally than in other settings. This result reinforces that the V\&L model forgot part of the knowledge from natural language through additional learning, and images could not fully complement them.

\subsection{Image Generation}

\paragraph{Settings} Similarly to the table generation, we chose OFA for the comparison. We additionally join the reference tables (Gold) and those generated by models in \S\ref{subsec:exp:table} (OFA, BART) as the input in order to investigate the impact of the ability to infer table knowledge. We also used the \texttt{base} settings with the hyperparameters reported in \citet{pmlr-v162-wang22al}. We also performed the training three times with different seeds and reported their average scores with their standard deviations\footnote{See Appendix \ref{appendix:exp:image} for the detailed settings.}.

\paragraph{Results} Table \ref{tab:results_image_gen} shows the results for each setting in the image generation\footnote{Appendix \ref{appendix:generated:images} shows the generated images.}. Since the CLIP value in OFA is close to the result \cite{pmlr-v162-wang22al} in MS COCO \cite{https://doi.org/10.48550/arxiv.1504.00325} for image generation, the use of our created dataset is reasonable for training models. In addition, the input of Table (Gold) improves all metrics, indicating that the model produces higher quality images when provided with complementary knowledge about the entities. This result also indicates that OFA does not retain sufficient knowledge of the entities in English Wikipedia.

In addition, we did not observe any performance improvement in CLIP and FID when fed with automatically generated tables from BART and OFA. However, tables generated by BART improves IS with the lower performance degradation of FID than that by OFA, indicating that automatically generated tables can improve the diversity of the output images and accurate tables are more important for improving performance in image generation.

\begin{table}[t]
\centering
\resizebox{\columnwidth}{!}{
\begin{tabular}{lccc}
\toprule
\textbf{Input}                    & \textbf{CLIP}~$\uparrow$ & \textbf{IS}~$\uparrow$   & \textbf{FID}~$\downarrow$  \\
\midrule
Title \& Caption                  & 28.7${\scriptstyle\pm0.0}$ & 10.5${\scriptstyle\pm0.1}$ & 31.1${\scriptstyle\pm0.2}$ \\
+Table (Gold) & \underline{\textbf{29.4}}${\scriptstyle\pm0.0}$ & \underline{\textbf{11.3}}${\scriptstyle\pm0.2}$ & \underline{\textbf{28.5}}${\scriptstyle\pm0.3}$  \\
+Table (BART) & 28.1${\scriptstyle\pm0.0}$ & 10.6${\scriptstyle\pm0.2}$ & 32.4${\scriptstyle\pm0.3}$ \\
+Table (OFA)  & 28.0${\scriptstyle\pm0.1}$ & 10.6${\scriptstyle\pm0.2}$ & 33.1${\scriptstyle\pm0.4}$ \\
\bottomrule
\end{tabular}
}
\caption{Image generation results. $\downarrow$ denotes that the lower the score, the more optimal the result. + denotes additionally used input to the title and captions. The parenthesis denotes the origin of the table. Other notations are the same as in Table \ref{tab:results_table_gen}.}
\label{tab:results_image_gen}
\end{table}

\section{Related Work}

Following the advancements in V\&L models \cite{ijcai2022p0762}, there have been various studies that investigate V\&L models. \citet{10.1007/978-3-030-58539-6_34} conducted a comprehensive analysis of V\&L models including the difference between model structures. Through their analysis, they revealed the importance of text information in V\&L tasks over image information. 

Several studies focused on the performance differences between V\&L models and text-only models. \citet{yun-etal-2021-vision-language} investigated the improvement of linguistic representations by pre-training V\&L models on PhysicalQA (PIQA) \cite{Bisk_Zellers_Le_bras_Gao_Choi_2020} and the probing framework of \cite{tenney2018what}. They concluded that the benefit of pre-trained V\&L models for text-only tasks is marginal. \citet{iki-aizawa-2021-effect,hagstrom-johansson-2022-adapt} compared the performance of V\&L models and text-only models on the text-only benchmark, GLUE \cite{wang-etal-2018-glue} and determined that the text-only model achieved higher scores than the V\&L models. 

However, even though various kinds of V\&L models \cite{NEURIPS2019_c74d97b0,Su2020VL-BERT:,Li_Duan_Fang_Gong_Jiang_2020,pmlr-v139-cho21a,pmlr-v162-wang22al,saharia2022photorealistic} inherit language-related knowledge from pre-trained language-only models, how the knowledge is inherited has yet to be investigated. Our work clarifies this by using our created dataset, Wikipedia Table and Image Generation (WikiTIG).

\section{Conclusion}

This paper investigates how knowledge about entities are preserved in a pre-trained V\&L model which is originally transferred from a pre-trained natural language model.

We analyzed a pre-trained V\&L model by creating the Wikipedia Table and Image Generation (WikiTIG) dataset for generating images and tables of the infoboxes in Wikipedia. WikiTIG consists of 200,000 infoboxes and their corresponding images from English Wikipedia.

Experimental results on a pre-trained V\&L model OFA \cite{pmlr-v162-wang22al} showed that the model forgot part of the knowledge about entities during pre-training, and the image information did not fully compensate for the forgotten knowledge.

\section*{Limitations}

Regarding the Wikipedia articles used for creating our dataset Wikipedia Table and Image Generation (WikiTIG), some infoboxes may not follow the defined format and rules. This is because various users can freely edit infoboxes. Moreover, the HTML dump data published by English Wikipedia is not based on recent information.

In image generation, due to the standard settings recommended by \citet{Zhang_2021_CVPR,pmlr-v139-ramesh21a,pmlr-v162-wang22al,10.1007/978-3-031-19787-1_41}, our image generation task requires generating a cropped fixed-size square image instead of the original aspect ratio.

In addition, a table in an infobox may contain cells unrelated to image generation, and thus it may be redundant for image generation.

\section*{Ethical Considerations}
In this study, we created our dataset from English Wikipedia. The editors of English Wikipedia remove unnecessarily offensive content and compile them into an encyclopedia (\url{https://en.wikipedia.org/wiki/Wikipedia:Offensive_material}). However, as stated on the official pages (\url{https://en.wikipedia.org/wiki/Wikipedia:Neutral_point_of_view#Bias_in_sources}, \url{https://en.wikipedia.org/wiki/Wikipedia:Reliable_sources#Biased_or_opinionated_sources}), the current English Wikipedia permits the use of biased information sources. Thus, there is a possibility that our created dataset also inherits the original biases of English Wikipedia.

\section*{Acknowledgments}

This work was supported by JSPS KAKENHI Grant Numbers JP21K17801, JP23H03458.

\bibliography{custom}
\bibliographystyle{acl_natbib}
\clearpage
\begin{table*}[!h]
\centering
\small
\begin{tabular}{lll}
\toprule
\textbf{Modality}     & \textbf{Task}                       & \textbf{Dataset}                                               \\
\midrule
Vision \& Language & Image Captioning          & \multirow{2}{*}{CC12M, CC3M, SBU, COCO, VG-Cap}      \\
       & Image-Text Matching       &                                                      \\
       \cmidrule{2-3} 
       & Visual Question Answering & VQAv2, VG-QA, GQA                                    \\
       \cmidrule{2-3}
       & Visual Grounding          & \multirow{2}{*}{RefCOCO, RefCOCO+, RefCOCOg, VG-Cap} \\
       & Grounded Captioning       &                                                      \\
       \midrule
Vision     & Detection                 & OpenImages, Object365, VG, COCO                      \\
       & Image Infilling           & OpenImages, YFCC100M, ImageNet-21K                   \\
       \midrule
Language     & Masked Language Modeling  & Pile \\                 
\bottomrule
\end{tabular}
\caption{Datasets used for pre-training OFA.}
\label{tab:ofa:dataset}
\end{table*}
\begin{table}[t]
\centering
\small
\begin{tabular}{lllll}
\toprule
\multicolumn{5}{c}{\textbf{Type Frequency}}         \\
\midrule
\textbf{Type}   & \textbf{Total}   & \textbf{Train}   & \textbf{Valid} & \textbf{Test}  \\
\midrule
Header & 12,804 & 12,071 & 3,373 & 3,401 \\
Group  & 201,937 & 183,728 & 13,252 & 13,444 \\
Value  & 772,392 & 705,556 & 54,292 & 55,162 \\
\midrule
\multicolumn{5}{c}{\textbf{Appearance Frequency}}   \\
\midrule
\textbf{Type}   & \textbf{Total}   & \textbf{Train}   & \textbf{Valid} & \textbf{Test}  \\
\midrule
Header & 1,535,791 & 1,383,138 & 75,870 & 76,783 \\
Group  & 518,125 & 466,337 & 25,745 & 26,043 \\
Value  & 1,535,791 & 1,383,138 & 75,870 & 76,783 \\
\bottomrule
\end{tabular}
\caption{Frequencies for each type of cells in each data split.}
\label{tab:freqs}
\end{table}
\begin{table}[t]
\centering
\small
\begin{tabular}{lllll}
\toprule
\multicolumn{5}{p{0.7\columnwidth}}{\textbf{Type frequencies of values for each header}}       \\
\midrule
          \textbf{Split} & \textbf{Mean}    & \textbf{Std.}     & \textbf{Max}     & \textbf{Min}    \\
           \midrule
All        & 60.3    & 548.4    & 18,518    & 1       \\
Train      & 58.5    & 516.5    & 17,050    & 1       \\
Valid      & 16.1    & 79.1     & 1,506     & 1       \\
Test       & 16.2    & 80.3     & 1,557     & 1       \\
\midrule
\multicolumn{5}{p{0.7\columnwidth}}{\textbf{Appearance frequencies of values for each header}} \\
\midrule
          \textbf{Split} & \textbf{Mean}    & \textbf{Std.}     & \textbf{Max}     & \textbf{Min}    \\
           \midrule
All        & 119.9    & 1244.0    & 48,150    & 1       \\
Train      & 114.6    & 1153.2    & 43,350    & 1       \\
Valid      & 22.5    & 118.7     & 2,391     & 1       \\
Test       & 22.6    & 119.4     & 2,409     & 1 \\
\bottomrule
\end{tabular}
\caption{Statitics of frequencies for values in each header. Std. denotes standard deviation, Max and Min denote maximum and minimum frequencies, respectively.}
\label{tab:header_stats}
\end{table}
\begin{table}[t]
\centering
\small
\begin{tabular}{lllll}
\toprule
\textbf{Split}     & \textbf{Mean} & \textbf{Std.} & \textbf{Max} & \textbf{Min} \\
      \midrule
All   & 17.6 & 7.6  & 149 & 1   \\
Train & 17.6 & 7.6  & 149 & 1   \\
Valid & 17.6 & 7.6  & 99 & 1   \\
Test  & 17.5 & 7.6  & 68 & 1  \\
\bottomrule
\end{tabular}
\caption{Statistics for the number of cells in tables. The notations are the same as Table \ref{tab:header_stats}.}
\label{tab:table_length}
\end{table}
\begin{table*}[t]
    \centering
    \small
    \resizebox{\textwidth}{!}{
    \begin{tabular}{p{3cm}cp{4cm}p{4cm}p{4cm}}
        \toprule
        \textbf{Title} & \textbf{Image} & \textbf{BART} & \textbf{OFA (Title \& Image)} & \textbf{Reference} \\
        \midrule
        \multirow{9}{*}{Low Pike} &
        \multirow{9}{*}{\includegraphics[width=3cm]{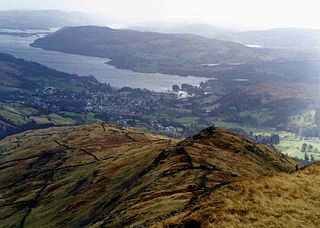}}
        & 
        Elevation | 1,859 m (3,927 ft) <> Location | England <> Range | Lake District <> Prominence | c. 1 m <> Parent peak | Low Pike <> Topo map | OS Landranger 89, 90, Explorer OL4 <> OS grid reference | NN93722 <> Listing | Marilyn, Hewitt, Nuttall &
        Elevation | 1,000 m (1,000 ft) <> Location | South England <> Coordinates | 45°49′0″N, 7°10′4″W <> Range | South East England <> > Range | south east england <> Topo map | CDT &
        Elevation | 508 m (1,667 ft) <> Range | Lake District, Eastern Fells <> Prominence | 28 m <> Parent peak | Dove Crag <> Topo map | OS Landranger 90 OS Explorer 7 <> OS grid reference | NY373077 <> Listing | Wainwright
       \\ \midrule
       \multirow{15}{*}{Ferruginous Pygmy-owl} &
        \multirow{15}{*}{\includegraphics[width=2.3cm]{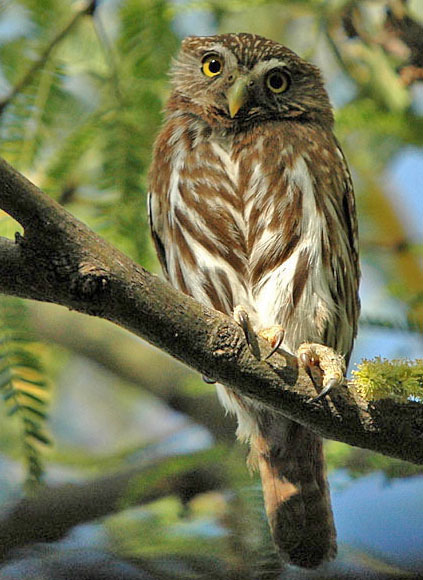}}
        & Conservation status <> Least Concern <> Scientific classification <> Kingdom: Animalia Phylum: Chordata Class: Aves Order: Passeriformes Family: Emberizidae Genus: Emberiza Species: E. ferruginus <> Domain: | Animalia
        &
        Conservation status <> Least Concern <> Scientific classification <> Kingdom: Animalia Phylum: Chordata Class: Aves Order: Passeriformes Family: Pterodactylidae Genus: Ferruginous Species: F. cinereus <> kingdom: | Animalia
        &
        Conservation status <> Least Concern <> Scientific classification <> Kingdom: Animalia Phylum: Chordata Class: Aves Order: Strigiformes Family: Strigidae Genus: Glaucidium Species: G. brasilianum <> Kingdom: | Animalia <> Phylum: | Chordata <> Class: | Aves <> Order: | Strigiformes <> Family: | Strigidae <> Genus: | Glaucidium <> Species: | G. brasilianum <> Binomial name <> Glaucidium brasilianum (Gmelin, 1788)
       \\ \midrule
       \multirow{13}{*}{Achlys (plant)} &
        \multirow{13}{*}{\includegraphics[width=2.3cm]{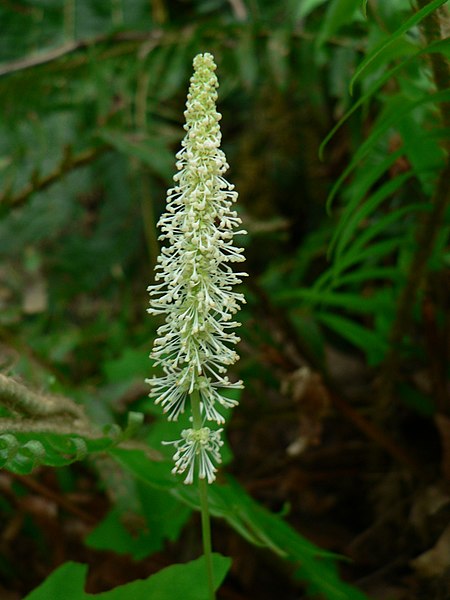}}
        &
        Scientific classification <> Kingdom: Plantae Division: Magnoliophyta Class: Liliopsida Order: Asparagales Family: Orchidaceae Subfamily: Higher Epidendroideae Genus: Achlys L.
        &
        Scientific classification <> Kingdom: Plantae Division: Magnoliophyta Class: Liliopsida Order: Asterales Family: Asteraceae Genus: Achlys Species: C. lilius <> kingdom: | Plantae <> Division: | Magnoliopsida <> Class: | Liliaceae <> Order: | Astrales <> Family: | Asteraceae
        &
        Scientific classification <> Kingdom: Plantae Division: Magnoliophyta Class: Magnoliopsida Order: Ranunculales Family: Berberidaceae Genus: Achlys DC. <> Kingdom: | Plantae <> Division: | Magnoliophyta <> Class: | Magnoliopsida <> Order: | Ranunculales <> Family: | Berberidaceae <> Genus: | Achlys DC. <> Species <> 2 or 3 - see text
       \\ \midrule
       \multirow{10}{*}{Giant’s Castle} &
       \multirow{10}{*}{\includegraphics[width=3cm]{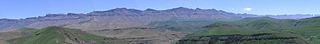}} 
       & 
       Developer(s) | Capcom <> Publisher(s): | Capcom (Japan) <> Platform(s: | PlayStation 2 <> Release date | JP November 15, 2002 NA November 20, 2002 <> Genre(s)/ | Adventure game <> Mode(es) | Single player, multiplayer <> Media | DVD-ROM <> Input methods | DualShock 2   Giant's Castle
       & 
       Elevation | 1,922 metres (1,923 ft) <> Location | New York, United States <> Coordinates | 41°44′00″N, 73°48′50″W <> Range | North York, New York <> Prominence | 2,944 metres (2,924 ft)
       & 
       Elevation | 3,315 metres (10,877 feet) <> Location | KwaZulu-Natal, South Africa <> Range | Drakensberg <> Coordinates | 29°20′S, 29°29′E <> Easiest route | scramble
       \\
        \bottomrule
    \end{tabular}}
    \caption{Tables generated by BART and OFA with title and image input and those of references. }
    \label{tab:gen_tables}
\end{table*}
\begin{table*}[t]
    \centering
    \small
    \resizebox{\textwidth}{!}{
    \begin{tabular}{p{3.8cm}ccc}
    \toprule
         \textbf{Input} & \textbf{w/ Tab.} & \textbf{w/o Tab.} & \textbf{Ref.}  \\
         \midrule
         \textbf{Title}: Upper Lake (Bhopal) & \multirow{9}{*}{\includegraphics[width=1.5cm]{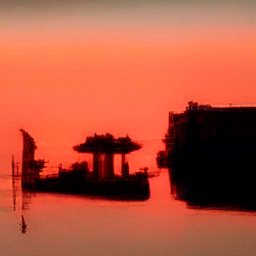}\includegraphics[width=1.5cm]{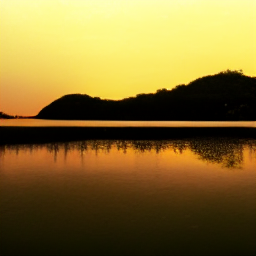}\includegraphics[width=1.5cm]{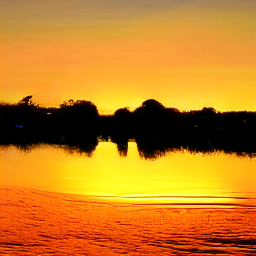}}& \multirow{9}{*}{\includegraphics[width=1.5cm]{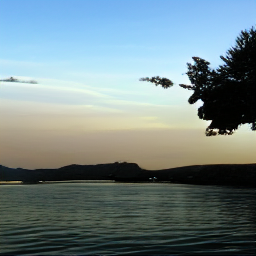}\includegraphics[width=1.5cm]{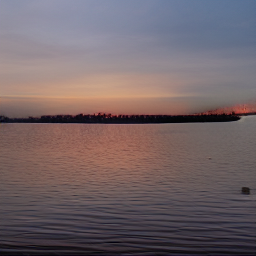}\includegraphics[width=1.5cm]{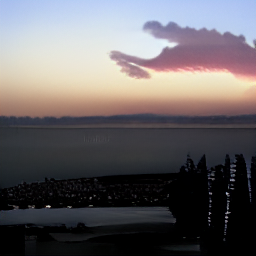}}& \multirow{9}{*}{\includegraphics[width=1.5cm]{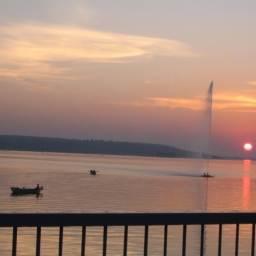}}\\
         \textbf{Caption}: Upper Lake (Bhopal) - Sunset & & & \\
         \textbf{Table}: Location | Madhya Pradesh, Bhopal <> Primary inflows | Kolans River <> Catchment area | 361 km² <> Basin countries | India <> Surface area | 31 km²
         &
         &
         &
         \\
         \midrule
         \textbf{Title}: May Lake & \multirow{11}{*}{\includegraphics[width=1.5cm]{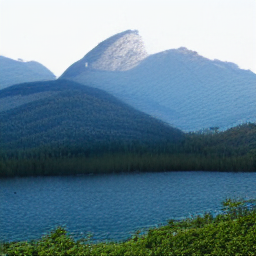}\includegraphics[width=1.5cm]{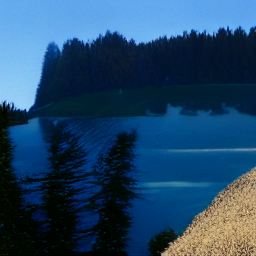}\includegraphics[width=1.5cm]{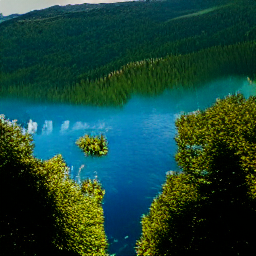}}& \multirow{11}{*}{\includegraphics[width=1.5cm]{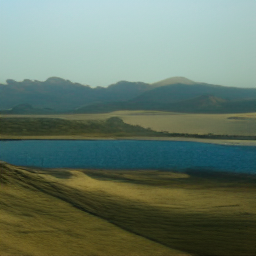}\includegraphics[width=1.5cm]{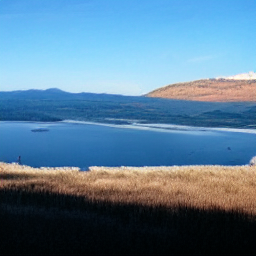}\includegraphics[width=1.5cm]{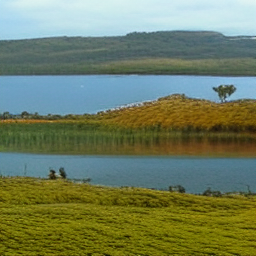}}& \multirow{11}{*}{\includegraphics[width=1.5cm]{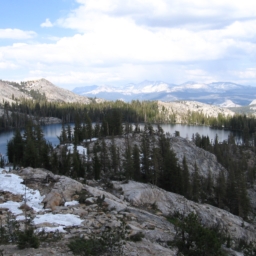}}\\
         \textbf{Caption}: May Lake - View from the trail up Mt. Hoffman. & & & \\
         \textbf{Table}: Location | Yosemite National Park, California <> Coordinates | 37°50′50″N, 119°29′37″WCoordinates: 37°50′50″N, 119°29′37″W <> Basin countries | United States <> Surface elevation | 9,270 ft (2,830 m)
         &
         &
         &
         \\
         \midrule
         \textbf{Title}: Littoral Rock-thrush & \multirow{20}{*}{\includegraphics[width=1.5cm]{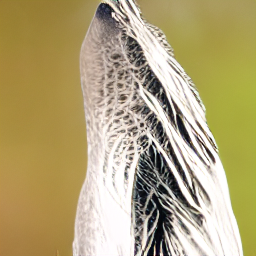}\includegraphics[width=1.5cm]{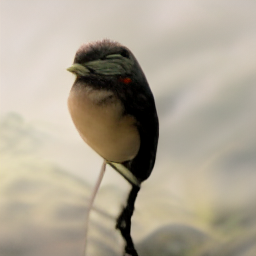}\includegraphics[width=1.5cm]{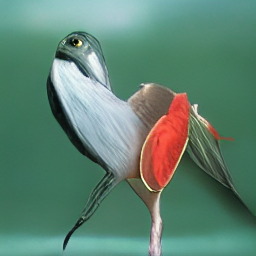}}& \multirow{20}{*}{\includegraphics[width=1.5cm]{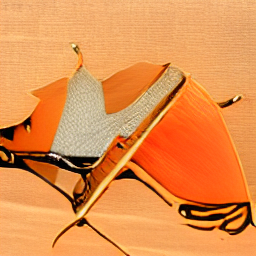}\includegraphics[width=1.5cm]{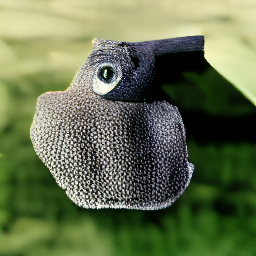}\includegraphics[width=1.5cm]{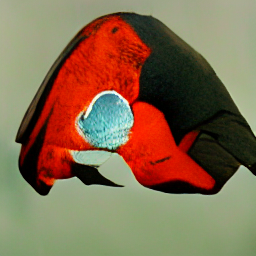}}& \multirow{20}{*}{\includegraphics[width=1.5cm]{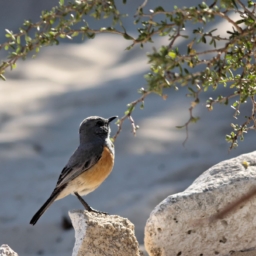}}\\
         \textbf{Caption}: Littoral Rock-thrush, M. imerinus & & & \\
         \textbf{Table}: Conservation status <> Least Concern <> Scientific classification <> Kingdom: Animalia Phylum: Chordata Class: Aves Order: Passeriformes Family: Muscicapidae Genus: Monticola Species: M. imerinus <> Kingdom: | Animalia <> Phylum: | Chordata <> Class: | Aves <> Order: | Passeriformes <> Family: | Muscicapidae <> Genus: | Monticola <> Species: | M. imerinus <> Binomial name <> Monticola imerinus (Hartlaub, 1860, St Augustine Bay, southeast Madagascar)
         &
         &
         &\\
         \midrule
         \textbf{Title}: Gießen (region) & \multirow{10}{*}{\includegraphics[width=1.5cm]{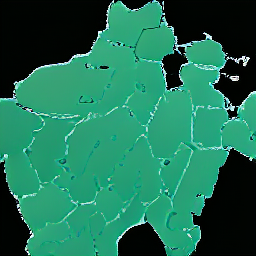}\includegraphics[width=1.5cm]{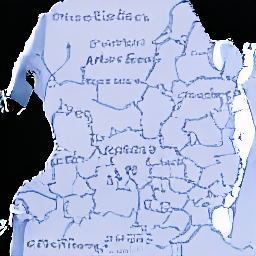}\includegraphics[width=1.5cm]{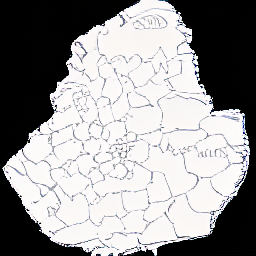}}& \multirow{10}{*}{\includegraphics[width=1.5cm]{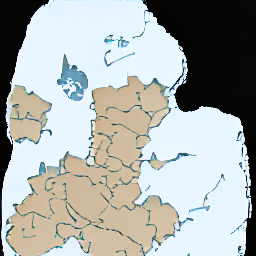}\includegraphics[width=1.5cm]{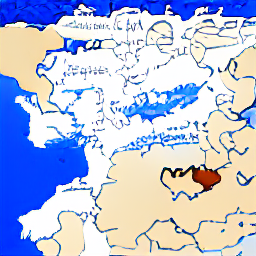}\includegraphics[width=1.5cm]{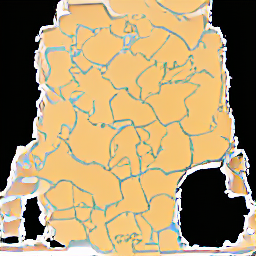}}& \multirow{10}{*}{\includegraphics[width=1.5cm]{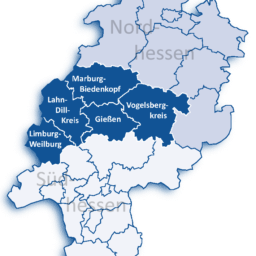}}\\
         \textbf{Caption}: Map of Hesse highlighting the Regierungsbezirk of Gießen & & & \\
         \textbf{Table}: State | Hesse <> District seat | Gießen <> Area | 5,381.14 km² <> Population | 1,061,444 (30 Sep. 2005) <> Pop. density | 197 /km² <> Web page | www.rp-giessen.de
         &
         &
         &
         \\
         \midrule
         \textbf{Title}: Giant's Castle & \multirow{9}{*}{\includegraphics[width=1.5cm]{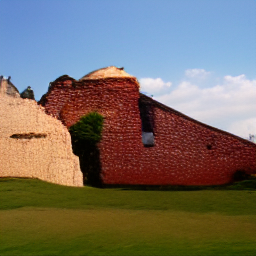}\includegraphics[width=1.5cm]{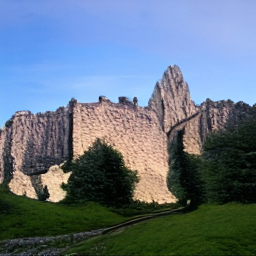}\includegraphics[width=1.5cm]{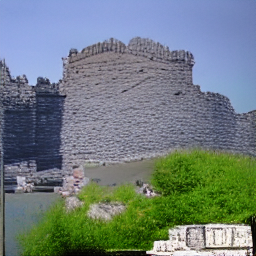}}& \multirow{9}{*}{\includegraphics[width=1.5cm]{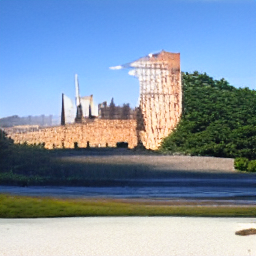}\includegraphics[width=1.5cm]{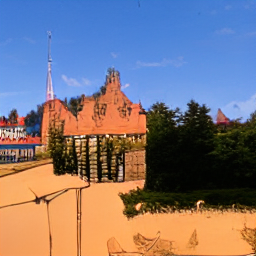}\includegraphics[width=1.5cm]{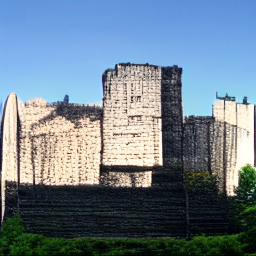}}& \multirow{9}{*}{\includegraphics[width=1.5cm]{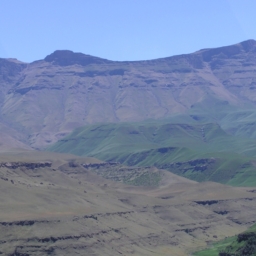}}\\
         \textbf{Caption}: Panorama at Giant's Castle & & & \\
         \textbf{Table}: Elevation | 3,315 metres (10,877 feet) <> Location | KwaZulu-Natal, South Africa <> Range | Drakensberg <> Coordinates | 29°20′S, 29°29′E <> Easiest route | scramble
         &
         &
         &
         \\
         
         \bottomrule
    \end{tabular}}
    \caption{Generated images. w/ Tab denotes the setting with tables, w/o Tab denotes the setting without tables, and Ref. denotes the reference images.}
    \label{tab:generated_images}
\end{table*}

\appendix

\section{Details of the datasets for pre-training OFA}
\label{appendix:ofa}

OFA pre-training uses various datasets for pre-training tasks in language, vision, and vision \& language modalities, as shown in Table \ref{tab:ofa:dataset}.
Note that 1.53\% of Pile \cite{https://doi.org/10.48550/arxiv.2101.00027} listed in Table \ref{tab:ofa:dataset} contains information from English Wikipedia. Therefore, we can understand that although OFA's pre-training focuses on V\&L tasks, it is also designed to prevent the knowledge acquired from natural language data from forgetting.

\section{Details of the metric calculation}
\subsection{Table-F$_1$}
\label{appendix:metric:tf1}
Let $e$ be an element of a target cell type. Here, we define a function $Match_{r,g}(e)$ that calculates the exact match of elements in reference and generated tables as follows:
\begin{align}
    & Match_{r,g}(e) \nonumber\\
    = & Min(Count_{r}(e),Count_{g}(e)),
\end{align}
where $Count_{r}(e)$, $Count_{g}(e)$ are functions that return frequencies of $e$ in a generated table $g$ and a reference table $r$, respectively. Note that $Min$ is a function that returns the minimum value from the given one. By using $Match_{r,g}(e)$, we calculate $Table\mathchar`-F_1$ as follows:
\begin{align}
    P(g, r) &= \frac{\sum_{e \in g} Match_{r,g}(e)}{\sum_{e' \in g} Count_{g}(e')}, \\
    R(g, r) &= \frac{\sum_{e \in r} Match_{r,g}(e)}{\sum_{e' \in g} Count_{r}(e')}, \\
    F_{1}(g, r) & = \frac{2P(g, r)R(g, r)}{P(g, r)+R(g, r)},
\end{align}
\begin{equation}
    Table\mathchar`-F_1 = \frac{1}{|D|}\sum_{(g,r) \in (G, R)}F_{1}(g, r),
\end{equation}
where $|D|$ denotes a number of tables, $G$ denotes all generated tables, and $R$ denotes all reference tables.

\subsection{Corpus-F$_1$}
\label{appendix:metric:cf1}
Instead of $Match_{r,g}(e)$, we define $Match_{R,G}(e)$ as follows: 
\begin{align}
    & Match_{R,G}(e) \nonumber\\
    = & Min(Count_{R}(e),Count_{G}(e)),
\end{align}
where $Count_{R}(e)$, $Count_{R}(e)$ are functions that return frequencies of $e$ in all generated tables $G$ and all reference tables $R$, respectively.
By using $Match_{R,G}(e)$, we calculate $Corpus\mathchar`-F_1$ as follows: 
\begin{align}
    P(G, R) &= \frac{\sum_{e \in G} Match_{R,G}(e)}{\sum_{e' \in G} Count_{G}(e')}, \\
    R(G, R) &= \frac{\sum_{e \in R} Match_{R,G}(e)}{\sum_{e' \in R} Count_{R}(e')},
\end{align}
\begin{equation}
    Corpus\mathchar`-F_1 = \frac{2P(G, R)R(G, R)}{P(G, R)+R(G, R)}.
\end{equation}

\section{Groups/Headers/Values in an infobox}
\label{appendix:group_names}

\begin{figure}[h!]
    \centering
    \includegraphics[width=0.7\columnwidth]{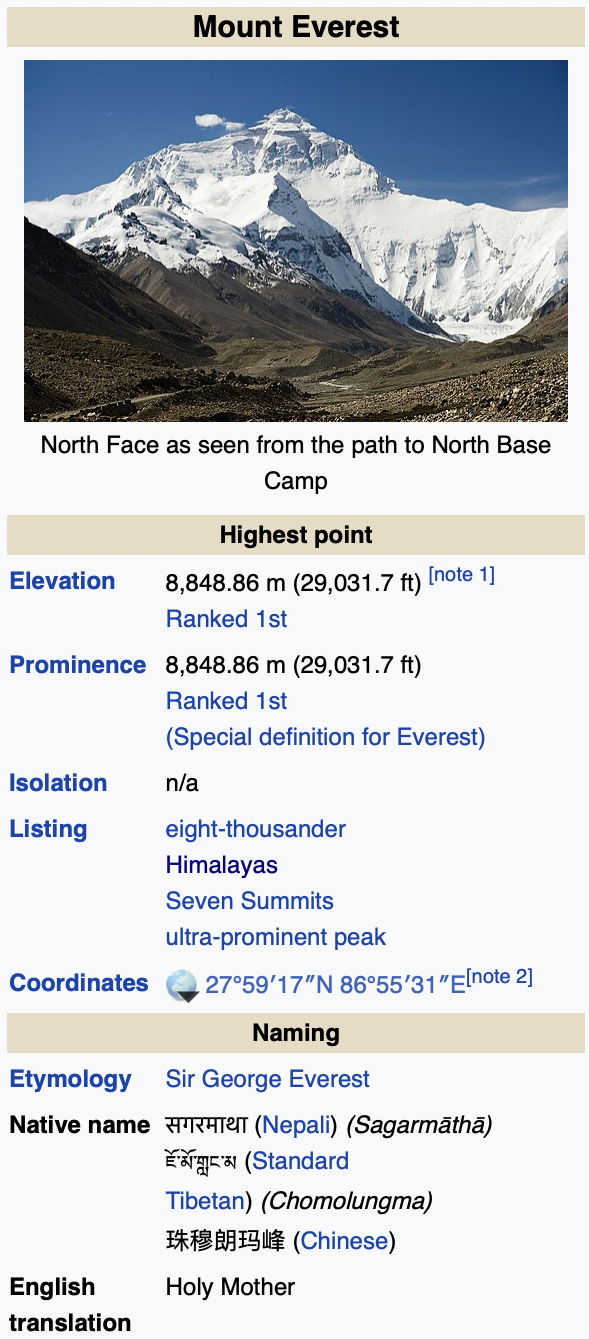}
    \caption{An example infobox with groups\footnotemark.}
    \label{fig:group_names}
\end{figure}
\footnotetext{\url{https://en.wikipedia.org/wiki/Mount_Everest}}

Figure \ref{fig:group_names} shows an example infobox that includes multiple groups. In this example, we can see two groups named with ``Highest point'' and ``Naming''.
The headers ``Elevation'', ``Prominence'', ``Isolation'', ``Listing'', and  ``Cordinates'' are grouped into ``Highest point''. The headers ``Etymology'', ``Native name'', and ``English translation'' are grouped into ``Naming''. The headers have corresponding values such as the value ``Holy Mother'' to the header ``English translation''. In the evaluation, we treat values as pairs with including their corresponding headers like (``English translation'', ``Holy Mother'') for the last row of the infobox in Figure \ref{fig:group_names}. 

\section{Details of our created dataset}
\label{appendix:dataset}

Wikipedia HTML dump data contains Wikipedia articles in HTML format, so we extracted infoboxes by using BeautifulSoup\footnote{\url{https://www.crummy.com/software/BeautifulSoup/bs4/doc/}}. Since the infoboxes contain links to the references of the main article in the form of [\#number], we removed them. We filtered out table rows that have more than two columns.

In table generation, if the short side of the input image exceeded 480px, we reduced the short side to 480px while maintaining the aspect ratio. In image generation, we changed the short side of the original image to 256px while maintaining the aspect ratio and then cropped the center of the image with a 256px square.

To measure the performance of both small and large models in the future, we also created additional datasets for the table generation with the short side of the image up to 256px and 384px, respectively.
Similarly, we also created a dataset for image generation with both sides of the image set to 128px.

For the sake of future expansion and to avoid data confusion, we divided the collected data into test data if the remainder of the SHA256 value of the title divided by 20 is 0, development data if the remainder is 1, and training data otherwise. Please see Table \ref{tab:stats} for the size of the dataset.

Table \ref{tab:freqs} shows the frequencies of each type of cells used for F$_1$ in \S\ref{subsec:tab_gen}. This result indicates that all types of cells have large number of type frequencies.

Table \ref{tab:header_stats} shows the statistics of frequencies for values in each header. Note that in Table \ref{tab:header_stats}, we do not take into account groups for the calculation different from the F$_1$ in \S\ref{subsec:tab_gen}.
From the table, we can understand that frequencies of values for each header have large variances.

Table \ref{tab:table_length} shows the statistics for the number of cells for each table. This result indicates that tables in infoboxes have the various number of cells.

Taking into account these results, we can understand that predicting cells based only on a label classification setting is difficult due to the various and diverse characteristics of the infobox tables.

To strictly comply with the license, we will only release text data to the public in the dataset release. For images, we will provide their URLs and preprocessing scripts for reproducing our dataset.

\section{Details of experimental settings}

For both tasks, we modified the publicly available implementation\footnote{\url{https://github.com/OFA-Sys/OFA} (Apache License 2.0).} by the authors of OFA. Since the released OFA uses the number of words after splitting by spaces for determining the maximum token length, we modified the OFA to use subwords to specify the maximum token length in the same way as BART. We set the maximum length for input and output in table and image generation to 1024 subwords. In addition, from the perspective of investigating the characteristics of the model and dataset, we used only maximum likelihood estimation for training and did not perform reinforcement learning. We ran training of each model three times with different seeds 0, 1, and 2.

\subsection{Table Generation}
\label{appendix:exp:table}

To avoid an unfair comparison of BART and OFA due to different implementations, we transferred BART's weight parameters\footnote{\url{https://dl.fbaipublicfiles.com/fairseq/models/bart.base.tar.gz} (MIT License).} to OFA and ran BART on OFA. We used the hyperparameters in the summarization of OFA for generation from titles. We also used the hyperparameters in captioning of OFA for generation from images. For a fair comparison, we used the captioning settings for all inferences. When the input includes titles, we used the prompt \textit{What is the infobox of " \{ENTITY\_NAME\} "?}. When the input only includes images, we used the prompt \textit{What is the infobox of the image?}.
We performed the text-only experiments with four RTX 3090s in one day and the image-included experiments with four RTX A6000s in one day.

\subsection{Image Generation}
\label{appendix:exp:image}

Basically, we inherited the hyperparameters used in OFA, but due to learning time, we set the beam size to 1 when generating images in the development data after each epoch in training. We used beam size 24 for testing, the same as in the original setting. We used the prompt \textit{What is the complete image? Caption: \{CAPTION\}} to generate images. When using tables, we combined the input with the delimiter \verb|<|\verb|>| at the end of the original input. We performed each experiment with four RTX A6000s in two days.

\section{Generated examples}
\subsection{Tables}
\label{appendix:generated:tables}

Table \ref{tab:gen_tables} shows the generated tables in the test data. In the first row regarding ``Low Pike'', BART generated a table for the mountain, whereas OFA generated a table for a city in the United Kingdom. This result is along with the result of the automatic evaluation that BART's prediction performance of values is better than other methods. However, even BART did not specify the detailed location of the mountain. This result indicates the difficulty of storing large amounts of geographic information in a pre-trained model.

In the second row regarding ``Ferruginous Pygmy-owl'', BART wrongly recognized it as a bunting (``Emberizidae''), at least a bird, and OFA wrongly recognized it as a pterosaur (``Pterodactylidae'').
Thus, this is a case that the forgotten knowledge about the entity was not completed with the image.

In the third row regarding ``Achlys (plant)'', both models recognized it as a plant (``Plantae''), and OFA precisely predicted its division as ``Magnoliopsida'' by the image. However, both models could not predict further details. This result indicates the difficulty of identifying plants with diverse species.

In the fourth row regarding ``Giant's Castle'', BART wrongly recognized it as a video game by its misleading name, even though OFA at least recognized it as a building in New York. The result is a case that the image supports the table generation by completing the knowledge about the entity. However, this support is not enough to generate precise information.

\subsection{Images}
\label{appendix:generated:images}

Table \ref{tab:generated_images} shows the generated images in the test data. In the first row, regarding ``Upper Lake (Bhopal)'', we can see both settings generated images along with the caption. Since such landscape photographs do not require the depiction of details, it is clear that images can be generated without detailed knowledge. 

In the second row regarding ``May Lake'', only w/ Tab. generated a lake with the mountain corresponding to the information in the table that shows the lake is at a high place. This result indicates that the table information can support generating images based on correct knowledge.

In the third row regarding ``Littoral Rock-thrush'', we can see that both w/ Tab. and w/o Tab. struggled to generate bird images. However, even in this difficult situation, w/ Tab. generated a more precise image than w/o Tab. by using the table information. This result is along with our automatic evaluation results that table information can improve image generation performances.

In the fourth row regarding ``Gießen (region)'', we can understand from this result that using a table alone is insufficient to generate precise images of geographic information.

We can see interesting results in the fifth row regarding ``Giant's Castle'', which is a mountain. Both w/o Tab. and w/ Tab. wrongly generated large castles due to the misleading name ``Giant's Castle''. Furthermore, w/ Tab. generated a large castle that looks like a mountain based on the knowledge of 3,315 meters in the table. This result indicates a limit to disambiguation based solely on the table.

\end{document}